\documentclass{article}

\PassOptionsToPackage{numbers, sort&compress}{natbib}

\usepackage[preprint]{neurips_2022}

\usepackage{graphicx}
\usepackage{amsmath}
\usepackage{amssymb}
\usepackage{booktabs}

\usepackage[pagebackref,breaklinks,colorlinks]{hyperref}

\usepackage[capitalize]{cleveref}
\crefname{section}{Sec.}{Secs.}
\Crefname{section}{Section}{Sections}
\Crefname{table}{Table}{Tables}
\crefname{table}{Tab.}{Tabs.}

\begin{document}

\title{Cancer-Net BCa-S: Breast Cancer Grade Prediction using Volumetric Deep Radiomic Features from Synthetic Correlated Diffusion Imaging}

\author{%
  Chi-en Amy Tai
  \qquad Hayden Gunraj
  \qquad Alexander Wong \\
  Vision and Image Processing Lab, University of Waterloo\\
  {\tt\small \{amy.tai, hayden.gunraj, alexander.wong\}@uwaterloo.ca}
}

\maketitle

\begin{abstract}
The prevalence of breast cancer continues to grow, affecting about 300,000 females in the United States in 2023. However, there are different levels of severity of breast cancer requiring different treatment strategies, and hence, grading breast cancer has become a vital component of breast cancer diagnosis and treatment planning. Specifically, the gold-standard Scarff-Bloom-Richardson (SBR) grade has been shown to consistently indicate a patient's response to chemotherapy. Unfortunately, the current method to determine the SBR grade requires removal of some cancer cells from the patient which can lead to stress and discomfort along with costly expenses. In this paper, we study the efficacy of deep learning for breast cancer grading based on synthetic correlated diffusion (CDI\textsuperscript{s}) imaging, a new magnetic resonance imaging (MRI) modality and found that it achieves better performance on SBR grade prediction compared to those learnt using gold-standard imaging modalities. Hence, we introduce Cancer-Net BCa-S, a volumetric deep radiomics approach for predicting SBR grade based on volumetric CDI\textsuperscript{s} data. Given the promising results, this proposed method to identify the severity of the cancer would allow for better treatment decisions without the need for a biopsy. Cancer-Net BCa-S has been made publicly available as part of a global open-source initiative for advancing machine learning for cancer care.
\end{abstract}

\section{Introduction}
The prevalence of breast cancer continues to grow, affecting about 300,000 females in the United States in 2023 \cite{american-cancer-death}. However, not all breast cancer is fatal. Specifically, the two main types of breast cancer are in situ and invasive breast cancer \cite{bc-types}. The former is a less severe form of breast cancer that is a precursor to the latter type. The latter type, invasive breast cancer, represents approximately 80\% of diagnosed cases and signifies that the cancer can or has already spread into the nearby tissue areas \cite{american-cancer-death, cancer-stats}. 

\begin{figure}
  \centering
    \includegraphics[width=\linewidth]{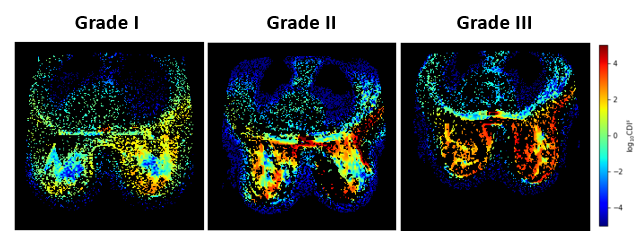}
  \caption{Example breast CDI\textsuperscript{s} images for the SBR grades.}
  \label{fig:grade-grid}
\end{figure}

Patients with invasive breast cancer also often receive a breast cancer grade that represents the similarity of the cancer cells to normal cells under the microscope. The three breast cancer grades (low, intermediate, and high) describe the speed of growth and likelihood of a good prognosis. Low grade (grade 1) cancer has the best prognosis with slow  growth and spread of the cancer, while high grade (grade 3) cancer has the worst prognosis with the greatest difference between cancer and normal cells and represent cancer that is fast-growing with quick spread to other cells. As such, the stage and grade of breast cancer are vital factors used to determine the severity of breast cancer and discern the best treatment strategy as the stage and grade have been shown to relate to the success of various treatment strategies \cite{breast-cancer-staging, cancer-stages, bc-grading, bc-grading2, grading-future}. Specifically, the gold-standard Scarff-Bloom-Richardson (SBR) grade (with example CDI\textsuperscript{s} shown in \cref{fig:grade-grid}) has been shown to consistently indicate a patient's response to chemotherapy \cite{sbr-useful, sbr-grade-sys}. 

\begin{figure*}
  \centering
    \includegraphics[width=\textwidth]{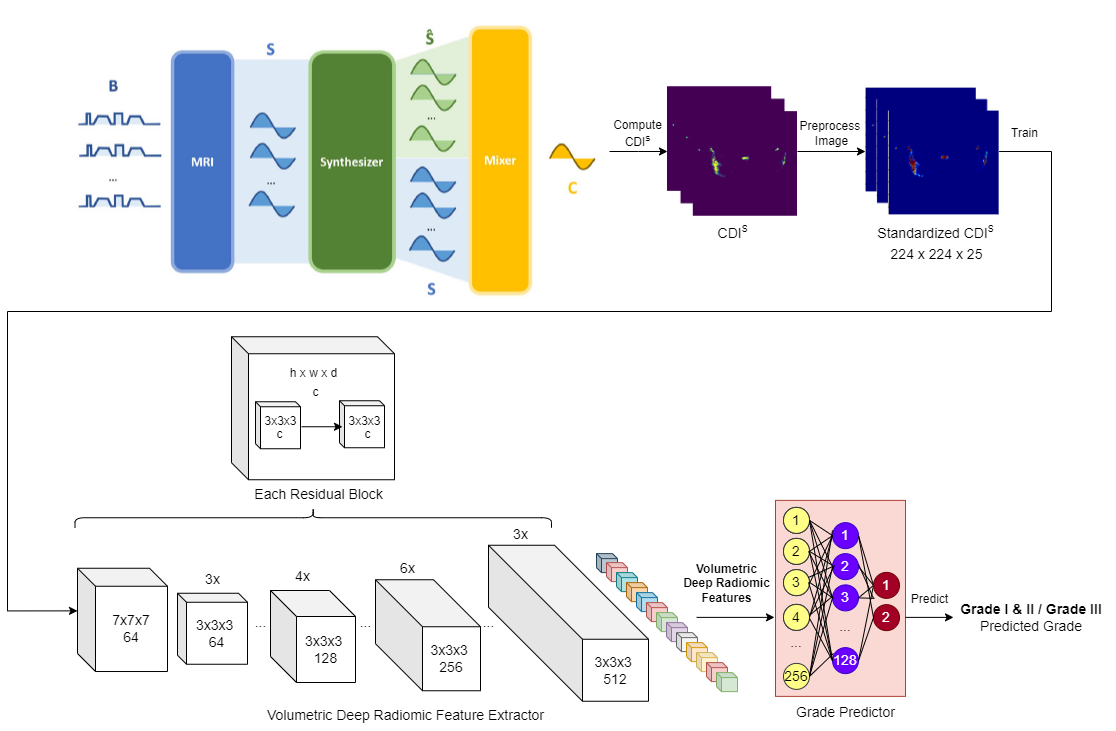}
  \caption{Adapted clinical support workflow for grade prediction using volumetric deep radiomic features from synthetic correlated diffusion imaging (CDI\textsuperscript{s}).}
  \label{fig:adapted-workflow}
\end{figure*}

Unfortunately, the current method to determine the SBR grade requires removal of some cancer cells from the patient which can lead to stress and discomfort along with costly expenses. In this paper, we study the efficacy of deep learning for breast cancer grading based on synthetic correlated diffusion (CDI\textsuperscript{s}) imaging, a new magnetic resonance imaging (MRI) modality and found that it achieves better performance on SBR grade prediction compared to those learnt using gold-standard imaging modalities. Hence, we introduce Cancer-Net BCa-S, a volumetric deep radiomics approach for predicting SBR grade based on volumetric CDI\textsuperscript{s} data. Given the promising results, this proposed method to identify the severity of the cancer would allow for better treatment decisions without the need for a biopsy. Cancer-Net BCa-S has been made publicly available \footnote{https://github.com/catai9/Cancer-Net-BCa} as part of a global open-source initiative for advancing machine learning for cancer care. The Cancer-Net BCa benchmark dataset that was used in this study is also publicly available  \footnote{https://www.kaggle.com/datasets/amytai/cancernet-bca} as a part of a global open-source initiative dedicated to accelerating advancement in machine learning to aid clinicians in the fight against cancer.

\section{Methodology}
The pre-treatment (T0) patient cohort of 252 patient cases across 10 different institutions obtained from the American College of Radiology Imaging Network (ACRIN) 6698/I-SPY2 study \cite{acrin6698-data-1, acrin6698-data-2, acrin6698-data-3, acrin6698-data-4} is used in this study. More specifically, CDI\textsuperscript{s} acquisitions was attained for each of the patient cases.  Furthermore, diffusion-weighted imaging (DWI) acquisitions, T2-weighted (T2w) acquisitions, and apparent diffusion coefficient (ADC) maps were also procured to compare against the performance of CDI\textsuperscript{s} for grade prediction. The SBR grade from the ACRIN 6698/I-SPY2 study was also obtained for learning and evaluation purposes to compare the current gold-standard MRI modalities and CDI\textsuperscript{s}. As seen in Table~\ref{sbr-grade-dist}, there is an uneven distribution of patients between the three grades and hence, SBR grade I and II were combined into one category. 

\begin{table}
 \caption{SBR grade distribution in the patient cohort.}
 \label{sbr-grade-dist}
 \centering
\begin{tabular}{ |c|c| } 
\hline
SBR Grade & Number of Patients \\
\hline
Grade I (Low) & 5 \\
Grade II (Intermediate) & 72 \\
Grade III (High) & 175 \\
 \hline
\end{tabular}
\end{table}

To investigate the efficacy of volumetric deep radiomic features from CDI\textsuperscript{s} for breast cancer grading, we adapt a previously introduced deep radiomic clinical support workflow~\cite{cancer-net-bca} to the proposed purpose of predicting the SBR grade of a patient (see \cref{fig:adapted-workflow}). First, we obtain CDI\textsuperscript{s} acquisitions for each patient by acquiring multiple native diffusion signals with different b-values, passing it into a signal synthesizer to produce synthetic signals and then mixing the native and synthetic signals together to obtain a final signal (CDI\textsuperscript{s}). To achieve dimensional consistency for machine learning, the CDI\textsuperscript{s} acquisitions are standardized into 224x224x25 volumetric data cubes. We then leverage a 34-layer volumetric residual convolutional neural network architecture trained on breast cancer image volumes~\cite{cancer-net-bca} to produce deep radiomic features for each patient based on their CDI\textsuperscript{s} data cubes. A grade predictor composed of a fully-connected neural network architecture is then learnt based on the extracted deep radiomic features from this volumetric network and categorized patient grade (Grade I/Grade II and Grade III). This grade predictor is used to predict the categorized patient grade prior to treatment. 

\section{Results and Discussion}
Leave-one-out cross-validation (LOOCV) is used to evaluate the efficacy of the proposed approach on the patient cohort with accuracy being the main performance metric of interest. The same 34-layer volumetric residual convolutional neural network architecture for volumetric deep radiomic feature extraction and same grade predictor architecture as described in Section 2 was used to evaluate the grade prediction efficiency for CDI\textsuperscript{s}as well as each gold-standard MRI modality (DWI, T2w, and ADC) for comparison consistency.

As seen in Table~\ref{grade-performance}, leveraging volumetric deep radiomic features for CDI\textsuperscript{s} achieves the highest grade predictive accuracy of 87.7\% with both sensitivity and specificity values over 80\%. Furthermore, CDI\textsuperscript{s} outperforms the gold-standard imaging modalities with an improvement of over 10\% on the next highest modality (T2w).

\begin{table}
 \caption{SBR grade prediction accuracy using LOOCV for different imaging modalities.}
 \label{grade-performance}
 \centering
\begin{tabular}{lrrr}
\hline
\textbf{Modality} & \multicolumn{1}{l}{\textbf{Accuracy}} & \multicolumn{1}{l}{\textbf{Sensitivity}} & \multicolumn{1}{l}{\textbf{Specificity}} \\
\hline
\textbf{CDIs} & \textbf{87.70\%} & \textbf{90.29\%} & \textbf{81.82\%} \\
T2w & 76.59\% & 99.43\% & 24.68\% \\
ADC & 69.44\% & 100.00\% & 0.00\% \\
DWI & 69.44\% & 95.43\% & 10.39\%
\end{tabular}
\end{table}

\section{Conclusion}
In this paper, we introduce Cancer-Net BCa-S, a volumetric deep radiomics approach for predicting SBR grade based on volumetric CDI\textsuperscript{s} data with a categorized grade prediction accuracy of 87.70\%. With the highest gold-standard MRI modality only achieving a prediction accuracy of 76.59\%, over 10\% lower than CDI\textsuperscript{s}, the proposed approach with CDI\textsuperscript{s} can increase the grade prediction performance compared to gold-standard MRI modalities. Given the promising results, future work involves expanding the study with a larger patient cohort to further validate our findings and leveraging improved CDI\textsuperscript{s} coefficient optimization to improve prediction performance.

\bibliographystyle{unsrtnat}
{
\small

\bibliography{egbib}
}

\end{document}